\documentclass[letterpaper, 10 pt, conference]{ieeeconf}
\IEEEoverridecommandlockouts
\pdfoutput=1
\let\chapter\section

\renewcommand\dbltextfloatsep{6pt}
\renewcommand\dblfloatsep{6pt}
\renewcommand\textfloatsep{6pt}
\usepackage{graphicx}
\usepackage{pifont}
\usepackage{cite}
\usepackage{booktabs}

\usepackage{subfig}
\usepackage[linesnumbered,ruled]{algorithm2e}
\usepackage[format=plain,labelsep=period,font=footnotesize]{caption}
\usepackage[bbgreekl]{mathbbol}
\usepackage{amsfonts}

\DeclareSymbolFontAlphabet{\mathbb}{AMSb}
\DeclareSymbolFontAlphabet{\mathbbl}{bbold}

\usepackage{amsmath,amsthm}
\usepackage{amssymb,mathrsfs}
\usepackage{array}
\usepackage{hyperref,color}
\usepackage{indentfirst}
\newtheorem{theorem}{Theorem}
\newtheorem{coroll}{Corollary}

\newtheorem{lemma}{Lemma}
\newtheorem{problem}{Problem}

\makeindex             

\usepackage[figuresright]{rotating}
\newcommand{\compresslist}{%
\setlength{\itemsep}{1pt}%
\setlength{\parskip}{0pt}%
\setlength{\parsep}{0pt}%
}

\allowdisplaybreaks 

\begin{document}
\title{\LARGE \bf A Near-Optimal Separation Principle for Nonlinear Stochastic Systems Arising in Robotic Path Planning and Control$ ^{\tiny{*}} $}
\author
{
Mohammadhussein Rafieisakhaei$^{1}$, Suman Chakravorty$^{2}$ and P. R. Kumar$^{1}$
\thanks{*This material is based upon work partially supported by NSF under Contract Nos. CNS-1646449 and Science \& Technology Center Grant CCF-0939370, the U.S. Army Research Office under Contract No. W911NF-15-1-0279, and NPRP grant NPRP 8-1531-2-651 from the Qatar National Research Fund, a member of Qatar Foundation.}
\thanks{$^{1}$M. Rafieisakhaei and P. R. Kumar are with the Department of Electrical and Computer Engineering, and $^{2}$S. Chakravorty is with the Department of Aerospace Engineering, Texas A\&M University, College Station, Texas, 77840 USA.
        \{\tt\small mrafieis, schakrav, prk@tamu.edu\}}%
}
\maketitle
\thispagestyle{empty}
\pagestyle{empty}
\begin{abstract}
We consider nonlinear stochastic systems that arise in path planning and control of mobile robots. As is typical of almost all nonlinear stochastic systems, the optimally solving problem is intractable. We provide a design approach which yields a tractable design that is quantifiably near-optimal. We exhibit a ``separation'' principle under a small noise assumption consisting of the optimal open-loop design of nominal trajectory followed by an optimal feedback law to track this trajectory, which is different from the usual effort of separating estimation from control. As a corollary, we obtain a trajectory-optimized linear quadratic regulator design for stochastic nonlinear systems with Gaussian noise.
\end{abstract}

\section{Introduction}\label{sec:Introduction}
Practical systems are often subject to inaccuracies that we model as noise. Planning for a stochastic system requires attention to the noise structure, available models and noise levels. Many robotic systems, in particular, mobile aerial and ground robots, are equipped with noisy actuators that require feedback compensation or planning ahead for a policy that accounts for the random perturbations. Simply ignoring the noise and planning for the unperturbed equivalent of the stochastic system can yield crucial errors leading to the failure in reaching the end-goal, or cause the system to fall into unsafe states.

In a stochastic setting, the general problem of sequential decision-making is formulated as a Markov Decision Problem (MDP) \cite{Kumar-book-86, bertsekas1995dynamic}. The optimal solution of the stochastic control problem can be obtained iteratively by value or policy iteration methods to solve the Hamilton-Jacobi-Bellman equations \cite{bertsekas1995dynamic}. Except in special cases, such as in a linear Gaussian environment, this involves discretization of the underlying spaces \cite{kushner2013numerical}; an approach whose scalability faces the curse of dimensionality \cite{Bellman:1957}. As a result, they require a computation time that is provably exponential in the state dimension, in a real number based model of complexity, without any assumption that $ P \neq NP $ \cite{chow1989complexity}.

Many approaches have been proposed based on their tractability. Some rely on a separate design of the deterministic trajectory from the feedback policy. Model Predictive Control (MPC)-based methods \cite{mayne2015robust,mayne2014model}, robust formulations \cite{tsitsiklis2007computational,le2007robust}, and other designs that relate to the Pontryagin's Maximum Principle \cite{kopp1962pontryagin} are some of the methods that have been successfully used as surrogate design approaches. Another popular approach is utilizing Differential Dynamic Programing (DDP) \cite{jacobson1970differential} and DDP-based variations, such as the Stochastic DDP \cite{theodorou2010stochastic}, iLQR and iLQG \cite{todorov2005generalized}. These methods rely on local linearizations of the cost function and the dynamics to the second order and propose iterative methods that attempt to find ``locally-optimal" solutions in a tube around a nominal trajectory \cite{todorov2005generalized}. 

In this paper, we address the nonlinear stochastic control problem and propose an architecture under which the separate design of an optimal open-loop control sequence and a feedback policy is near-optimal. In particular, we show that under a small noise assumption, the separation into globally-optimal trajectory design and a globally-optimal feedback control law holds for a fully-observed nonlinear stochastic system. This result also sheds light on the conditions under which popular design approaches based on the Maximum Principle may be globally $ \epsilon $-optimal.

We quantify the first order stochastic error for small-noise levels based on Wentzell-Freidlin large-deviations theory. We thereby determine reach to a Trajectory-optimized Linear Quadratic Regulator (T-LQR) design for fully-observed nonlinear stochastic systems under Gaussian small-noise perturbations. In short, the design can be broken into two parts: \textit{i)} an open-loop optimal control problem that designs the nominal trajectory of the LQR controller, which respects the nonlinearities as well as state and control constraints; \textit{ii)} the design of an LQR policy around the optimized nominal trajectory. The quality of the design is rigorously provided by the main results of the paper.

The organization of the paper is as follows. Section \ref{sec:Small Random Perturbations of a Non-Linear System } provides a brief background on Wentzell-Freidlin theory \cite{Freidlin1984} and investigates its implications regarding the linearization of a stochastic system coupled with the usage of the Taylor theorem. Section \ref{sec:General Problem: Fully Observed System} defines a general stochastic control problem for a fully-observed system. Section \ref{sec:Separation of Open Loop and Closed Loop Designs: Fully Observed Systems} provides the main results by first analyzing the effect of feedback compensation on the linearization error, and then providing the state and control error propagations along with probabilistic bounds based on the theory developed in Section \ref{sec:Small Random Perturbations of a Non-Linear System }. Section \ref{sec:Separation of Open Loop and Closed Loop Designs: Fully Observed Systems} also provides the first-order expected error of the stochastic cost function along with the separation result. Section \ref{sec:T-LQR: Trajectory-optimized LQR} introduces the T-LQR design approach. Finally, Section \ref{sec:Example} provides a design based on T-LQR for a non-holonomic car-like robot and provides numerical results on the proposed approach to design.

\section{Small Random Perturbations of a Non-Linear System}\label{sec:Small Random Perturbations of a Non-Linear System }
In this section, we discuss the theoretical background regarding the small noise perturbations of general dynamical systems. In particular, we discuss Wentzell-Freidlin theory on the small noise asymptotics of a perturbed system represented by a general Stochastic Differential Equation (SDE). We consider a time-varying system as that is required for our design. A general discussion regarding large deviations of the trajectories of a perturbed system from that of its unperturbed counterparts and related theories can be found in \cite{Freidlin1984,wentzell2012limit,dembo2009large,fleming1971stochastic,Cruz-Suarez-stoc,Perkins1976,perkins1975nonlinear,OCAOCA4660020109,varadhan1984large}.

\textit{Probability space:} We consider a probability space $ \{\mathbbl{\Omega}, \mathscr{F}, P\} $ with the random variables on a measurable space $ (\mathbb{X}, \mathscr{B}) $, where $ \mathbb{X} $ is a Euclidean space with dimension of $ n_x $, $ n_{w} $ or a smooth manifold in these spaces, and $ \mathscr{B} $ is the corresponding $ \sigma $-algebra of Borel sets.

\textit{Diffusion process:} Let us consider a dynamical system with the following equation:
\begin{equation}\label{eq:general SDE time-varying}
d\mathbf{X}^{\epsilon}_{t}=\mathbf{b}(t, \mathbf{X}^{\epsilon}_{t})dt+\epsilon d\boldsymbol{w}_{t},~ \mathbf{{X}}^{\epsilon}_{0}=\mathbf{x}_{0},
\end{equation}
where $ \mathbf{b}:\mathbb{R}\times\mathbb{R}^{n_x}\rightarrow\mathbb{R}^{n_x} $ is a uniformly Lipschitz continuous function, such that:
\begin{align}\label{eq:Lipschitz continuous time-varying}
|\!|\mathbf{b}(t_1,\mathbf{x}_1)\!-\!\mathbf{b}(t_2,\mathbf{x}_2)|\!|\!\le\! K_1|\!|\mathbf{x}_1\!-\!\mathbf{x}_2|\!|,
\end{align}
where $ \mathbf{x}_1, \mathbf{x}_2\in \mathbb{R}^{n_x} $, $ t_1, t_2\in[0, K] $, $ \epsilon>0 $, and $ K_1>0 $, $ \{\boldsymbol{w}_{t}, t\ge 0\} $ is a Wiener process on $ \mathbb{R}^{n_{w}} $. 

\textit{Nominal unperturbed trajectory:} Such a system can result from small random perturbations of the following time-varying ODE:
\begin{equation}\label{eq:general ODE time-varying}
\mathbf{\dot{x}}^{p}_{t}=\mathbf{b}(t, \mathbf{x}^{p}_{t}),
\end{equation}
with initial condition $ \mathbf{x}^{p}_{0}=\mathbf{x}_{0}\in \mathbb{R}^{n_x} $.

\textit{First order Taylor expansion:} Using Taylor's theorem to obtain the first order linearization of the right hand side of the above system around the trajectory $ \{\mathbf{x}^{p}_{t}\}_{t=0}^{K} $ results in the following:
\begin{equation}\label{eq:general SDE non-additive linearized time-varying}
d\mathbf{{X}}^{\epsilon}_{t}\!=\!\mathbf{b}(t,\mathbf{x}^{p}_{t})dt\!+\!\mathbf{A}_{t}(\mathbf{X}^{\epsilon}_{t}-\mathbf{x}^{p}_{t})dt+\!\epsilon d\boldsymbol{w}_{t}+o(|\!|\mathbf{X}^{\epsilon}_{t}-\mathbf{x}^{p}_{t}|\!|),
\end{equation}
where $ \mathbf{A}_{t}=\nabla_{\mathbf{x}} \mathbf{b}(t,\mathbf{x})|_{t, \mathbf{x}^{p}_{t}} $ is the Jacobian matrix.

\textit{Accuracy of linearization:} Equation \eqref{eq:general SDE non-additive linearized time-varying} states that if $ |\!|\mathbf{X}^{\epsilon}_{t}-\mathbf{x}^{p}_{t}|\!|\le\delta $ for all $ 0\le t\le K $, then, 
\begin{equation}\label{eq:general SDE non-additive linearized time-varying_2}
d\mathbf{{X}}^{\epsilon}_{t}\!=\!\mathbf{b}(t,\mathbf{x}^{p}_{t})dt\!+\!\mathbf{A}_{t}(\mathbf{X}^{\epsilon}_{t}-\mathbf{x}^{p}_{t})dt+\!\epsilon d\boldsymbol{w}_{t}+o(\delta).
\end{equation}
We will use the Wentzell-Freidlin theorem to calculate the probability that the aforesaid condition holds. In order to do that, we define the action functional for the family of processes defined in equation \eqref{eq:general SDE time-varying}.

\textit{Action functional \cite{Freidlin1984}:} For $ [T_1, T_2]\subseteq[0, K] $, the action functional is defined as:
\begin{align}\label{eq:action functional time-varying}
S_{T_1,T_2}(\boldsymbol{\phi}):=\frac{1}{2\epsilon^{2}}\int_{T_1}^{T_2}|\!|\boldsymbol{\dot{\phi}}_{t}-\mathbf{b}(t,\boldsymbol{\phi}_t)|\!|^{2}dt,
\end{align}
for absolutely continuous $ \boldsymbol{\phi} $, and is set to be equal to $ +\infty $ for other $ \boldsymbol{\phi}\in \mathbb{C}_{0K}(\mathbb{R}^{n_{x}}) $. Note that this defines the action functional for the ($ \epsilon $-dependent) family of processes given by the SDE \eqref{eq:general SDE time-varying}, uniformly on the whole space as $ \epsilon\downarrow0 $.

\begin{theorem}\textup{\textbf{Exponential Rate of Convergence}}\label{theorem: Rate of Convergence } Let:
\begin{itemize}
\item $ \mathbb{D} $ be a domain in $ \mathbb{R}^{n_x} $, and denote its closure by $ \mathrm{cl}(\mathbb{D}) $;
\item $ \partial \mathbb{D} $ denote the boundary of $ \mathbb{D} $;
\item $
\mathbb{H}_{\mathbb{D}}(t, \mathbf{x}_{0})\!\!=\!\!\{\boldsymbol{\phi}\in\mathbb{C}_{0K}(\mathbb{R}^{n_x}):\boldsymbol{\phi}_{0}=\mathbf{x}_{0},\boldsymbol{\phi}_{t}\in\mathbb{D}\cup\partial\mathbb{D} \}.$ 
\end{itemize} 
Assume $ \partial\mathbb{D} = \partial\mathrm{cl}(\mathbb{D}) $. Then, we have the following:
\begin{align}
\lim\limits_{\epsilon\rightarrow 0}\epsilon^{2} \ln P_{\mathbf{x}_{0}}\{\mathbf{X}^{\epsilon}_{t}\in\mathbb{D} \}&\!\!=\!\!-\!\!\!\!\!\inf\limits_{\boldsymbol{\phi}\in\mathbb{H}_{\mathbb{D}}(t, \mathbf{x}_{0})}S_{0t}(\boldsymbol{\phi}), \label{eq:asymptotics D_t 1 }
\end{align}
\end{theorem}

\begin{theorem}\textup{\textbf{Asymptotics of the Diffusion Process:}} Let:
\begin{itemize}
\item $ \mathbb{D}_{t}=\mathrm{cl}(\mathbb{B}^{c}_{\delta}(\mathbf{x}^{p}_{t})) $, the closure of the complement of a ball with radius $ \delta>0 $ around the point $ \mathbf{x}^{p}_{t} $; and
\item $ \tau^{\epsilon}=\mathrm{Min}\{t:\mathbf{X}^{\epsilon}_{t}\in\mathbb{D}_{t} \} $.
\end{itemize} Then, 
\begin{align}
\lim_{\epsilon \to 0} \epsilon^2 \ln P_{\mathbf{x}_{0}}\{\tau^\epsilon \leq t\} 
= -\!\!\!\! \inf_{\{\boldsymbol{\phi}: \boldsymbol{\phi}_0 = \mathbf{x}_0, |\!|\boldsymbol{\phi}_t - \mathbf{x}_t^p|\!| > \delta \}} S_{0t}(\boldsymbol{\phi}).
\end{align}
\end{theorem}
Proof of these results can be found in \cite{Freidlin1984, wentzell2012limit}.

Thus, according to Theorem \ref{theorem: Rate of Convergence }, for a given $ t $, the probability as $ \epsilon\downarrow0 $ of $ |\!|\mathbf{X}^{\epsilon}_{t}-\mathbf{x}^{p}_{t}|\!|\ge\delta $ can be calculated as in equation \eqref{eq:asymptotics D_t 1 }. Note that this probability tends to zero exponentially for any fixed $ \delta>0 $ as $ \epsilon\downarrow0 $. Moreover, from Theorem 2, the probability that the trajectory of $ \mathbf{X}^{\epsilon} $ ever exits the tube of radius $\delta$ round the nominal trajectory in the time interval $ [0,t] $ also goes to zero exponentially at the same rate. (This also asserts that the likely paths to ever exit in $ [0,t] $ are those exiting at time $ t $). This provides the validity region of the linearized equation \eqref{eq:general SDE non-additive linearized time-varying} and concludes our discussion in this section. 

\section{The Fully Observed System}\label{sec:General Problem: Fully Observed System}
The general stochastic control problem of interest for fully observed system can be formulated as an optimization problem in the space of feedback policies. In this section, we define the system equations and pose the general problem. Without loss of generality, we consider the discrete-time version of the systems considered in the previous section and continue our analysis on that basis.

\emph{Process model:} We denote the state and control by $\mathbf{x}\in \mathbb{X} \subset\mathbb{R}^{n_x}$ and $\mathbf{u}\in \mathbb{U}\subset\mathbb{R}^{n_u}$, respectively. The process model with $\mathbf{f}:\mathbb{X}\times\mathbb{U}\rightarrow\mathbb{X}$ is defined as:  
\begin{equation}\label{eq:non-linear system equations fully observed}
\mathbf{x}_{t+1}=\mathbf{f}(\mathbf{x}_{t},\mathbf{u}_{t})+\boldsymbol{\omega}_{t}, ~~\boldsymbol{\omega}_{t}\sim \mathcal{N}(\mathbf{0}, \boldsymbol{\Sigma}_{\boldsymbol{\omega}_{t}}) 
\end{equation}
where $ \{\boldsymbol{\omega}_t\} $ is independent, identically distributed (i.i.d.).

Now, we pose the general stochastic control problem \cite{Kumar-book-86,Bertsekas07}.

\begin{problem}\label{problem:Stochastic Control Problem fully observed} \textup{\textbf{Stochastic Control Problem for Fully Observed System}}: Given an initial state $ \mathbf{x}_{0} $, we wish to determine an optimal or near-optimal for
\begin{align}\label{problem eq:Stochastic Control Problem fully observed}
\nonumber \min_{\pi}~&\mathbb{E}[\sum_{t=0}^{K-1}c_t^{\pi}(\mathbf{x}_t,\mathbf{u}_t)+c_K^{\pi}(\mathbf{x}_K)]
\\ s.t.&~\mathbf{x}_{t+1}=\mathbf{f}(\mathbf{x}_{t},\mathbf{u}_t)+\boldsymbol{\omega}_{t},
\end{align}
where the optimization is over Markov, i.e., time-varying state-feedback, policies, $ \pi\in\mathbbl{\Pi} $, with
\begin{itemize}\compresslist
\item $ \pi:=\{\pi_{0}, \cdots, \pi_{t}\} $, $ \pi_{t} :\mathbb{X}\rightarrow \mathbb{U} $ ;
\item and $ \mathbf{u}_{t}=\pi_{t}(\mathbf{x}_{t}) $ specifying the action taken given the state;
\item $ c^{\pi}_t(\cdot,\cdot):\mathbb{X}\times\mathbb{U}\rightarrow\mathbb{R} $ is the one-step cost function;
\item $ c_K^{\pi}(\cdot):\mathbb{X}\rightarrow\mathbb{R} $  denotes the terminal cost;
\item $ K $ is the time horizon.
\end{itemize}
\end{problem}

\section{Separation of Open Loop and Closed Loop Designs: Fully Observed Systems}\label{sec:Separation of Open Loop and Closed Loop Designs: Fully Observed Systems}
In this section, we provide the theoretical basis for our design. The analysis employs the Taylor series expansion of the process model and large deviations theory.

\subsection{Preliminary Analysis}
We start by providing the nominal trajectory to linearize the process model. Then, we discuss the feedback law and compensate the process model with the feedback in order to use large deviations theory.

\textit{Nominal Trajectory}: We use the process model with zero noise to propagate the initial state, $ \mathbf{x}_{0} $, with a set of unknown controls $ \{ \mathbf{u}^{p}_{t}\}_{t=0}^{K-1} $, in order to obtain a parametrization of the feasible nominal trajectories as:
\begin{align}\label{eq:p-traj}
\mathbf{x}^{p}_{t+1}=\mathbf{f}(\mathbf{x}^{p}_t, \mathbf{u}^{p}_t),~~ 0\le t \le K\!-\!1,
\end{align}
where $ \mathbf{x}^{p}_{0} = \mathbf{x}_{0} $.

\textit{Linearization of the process model:} We linearize the process model of equation \eqref{eq:non-linear system equations fully observed} around the nominal trajectory:
\begin{align}\label{eq:linearized system fully observed}
\tilde{\mathbf{x}}_{t+1}&\!=\!\mathbf{A}_t\tilde{\mathbf{x}}_t \!+\! \mathbf{B}_t\tilde{\mathbf{u}}_t \!+\!\boldsymbol{\omega}_t\!+\!o(e^{\mathbf{x},\mathbf{u}}_{t}),
\end{align}
where we have:
\begin{itemize}\compresslist
\item $ \mathbf{A}_t(\mathbf{x}^{p}_{t},\mathbf{u}^{p}_t)=\nabla_{\mathbf{x}} \mathbf{f}(\mathbf{x},\mathbf{u})|_{ \mathbf{x}^{p}_{t}, \mathbf{u}^{p}_{t}} $, denoted by $ \mathbf{A}_t $;
\item $ \mathbf{B}_t(\mathbf{x}^{p}_{t},\mathbf{u}^{p}_t)=\nabla_{\mathbf{u}} \mathbf{f}(\mathbf{x},\mathbf{u})|_{ \mathbf{x}^{p}_{t}, \mathbf{u}^{p}_{t}} $, denoted by $ \mathbf{B}_t $;
\item $ \tilde{\mathbf{x}}_{t}:=\mathbf{x}_t-\mathbf{x}^{p}_{t} $, the state error  with respect to the nominal trajectory;
\item $ \tilde{\mathbf{u}}_{t}:=\mathbf{u}_t-\mathbf{u}^{p}_{t} $, the control error; and
\item $ e^{\mathbf{x},\mathbf{u}}_{t}:=|\!|\tilde{\mathbf{x}}_t|\!|+|\!|\tilde{\mathbf{u}}_t|\!| $ the error.
\end{itemize}
As the control inputs change, the underlying nominal trajectory also changes, and therefore the Jacobian matrices, $ \mathbf{A}_t $, $ \mathbf{B}_t $, and $ \mathbf{G}_t $ change, as well. The Taylor series expansion of equation \eqref{eq:linearized system fully observed} is valid as $ e^{\mathbf{x}}_{t}\rightarrow 0 $, i.e., the linearized function remains close to the linearization region. In this equation, the only factor that can drive the linearized function away from the linearization region is the noise process $ \boldsymbol{\omega}_t $. Therefore, we establish probabilistic bounds on the validity of this equation using the small noise theory of Section \ref{sec:Small Random Perturbations of a Non-Linear System }.

\textit{Optimization over policy space:} A feedback law with Linear Time-Varying (LTV) gain is sufficient to control a linearized model around a nominal trajectory. Therefore, we restrict the search to feedback policies with LTV feedback gain, $ \mathbbl{\Pi}^{L} $. In the next section, we design a Linear Quadratic Regulator policy (LQR) as a special case for our design.

\textit{Feedback controller:} Assuming the controllability of the deterministic model of the system, we suppose the existence of a feedback control law with LTV feedback gain to track and stabilize the trajectory of states around the nominal-designed trajectory. Later, we explain in detail how to design such a law. Thus, the control action error can be expressed as:
\begin{align}\label{eq:feedback law}
\tilde{\mathbf{u}}_{t}=\mathbf{u}_t-\mathbf{u}^{p}_{t}=-\mathbf{L}_{t}(\mathbf{x}_{t}-\mathbf{x}^{p}_{t}),
\end{align}
where $ \mathbf{L}_{t} $ is the linear feedback gain. It is important to note that although we are working with the linearized system, the original system is a nonlinear system, and the design is tailored to work for the original system.

\textit{Linearized system equation compensated with feedback:} Replacing the feedback law in equation \eqref{eq:linearized system fully observed}, we obtain:
\begin{align}\label{eq:linearized system with feedback fully observed}
\nonumber\tilde{\mathbf{x}}_{t+1}=&\mathbf{A}_t\tilde{\mathbf{x}}_t + \mathbf{B}_t\tilde{\mathbf{u}}_t +\boldsymbol{\omega}_t+o(e^{\mathbf{x},\mathbf{u}}_{t}),
\\\nonumber=&(\mathbf{A}_t-\mathbf{B}_t\mathbf{L}_t)\tilde{\mathbf{x}}_t +\boldsymbol{\omega}_t+o(e^{\mathbf{x}}_{t}),
\\=&\mathbf{D}_t\tilde{\mathbf{x}}_t +\boldsymbol{\omega}_t+o(e^{\mathbf{x}}_{t}),
\end{align}
where $ \mathbf{D}_{t}:=\mathbf{A}_t- \mathbf{B}_t\mathbf{L}_t, t\ge 1 $ and $ e^{\mathbf{x},\boldsymbol{\omega}}_{t}:=|\!|\tilde{\mathbf{x}}_t|\!| $ denotes the linearization-based error.

\textit{Compensating the original system with feedback:} Let us substitute for the control action in \eqref{eq:non-linear system equations fully observed} using the feedback law of \eqref{eq:feedback law} as follows:
\begin{align}
\nonumber\mathbf{x}_{t+1}=\mathbf{f}(\mathbf{x}_{t},\mathbf{u}_{t})+\boldsymbol{\omega}_{t}=\mathbf{f}(\mathbf{x}_{t},\mathbf{u}^{p}_{t}-\mathbf{L}_{t}(\mathbf{x}_{t}-\mathbf{x}^{p}_{t}))+\boldsymbol{\omega}_{t}.
\end{align}
Using the last equation we define $ \mathbf{g}:\mathbb{R}\times\mathbb{X}\rightarrow\mathbb{X} $, where
\begin{align}
\mathbf{g}(t,\mathbf{x})=:\mathbf{f}(\mathbf{x}_{t},\mathbf{u}^{p}_{t}-\mathbf{L}_{t}(\mathbf{x}_{t}-\mathbf{x}^{p}_{t})).\label{eq:g function}
\end{align}
Note that the time-dependency for $ g $ stems from the time-dependency of the feedback law. Moreover, the nominal trajectory, $ \{\mathbf{x}^{p}_{t}\}_{t=0}^{K} $, satisfies the same equation as \eqref{eq:p-traj}:
\begin{align*}
\mathbf{x}^{p}_{t+1}&=\mathbf{g}(t,\mathbf{x}^{p}_{t})=\mathbf{f}(\mathbf{x}^{p}_{t},\mathbf{u}^{p}_{t}-\mathbf{L}_{t}(\mathbf{x}^{p}_{t}-\mathbf{x}^{p}_{t}))=\mathbf{f}(\mathbf{x}^{p}_{t},\mathbf{u}^{p}_{t}).
\end{align*}
Note that linearizing $ \mathbf{g} $ around the nominal trajectory yields \eqref{eq:linearized system with feedback fully observed}, which itself is equivalent to equation \eqref{eq:linearized system fully observed}
\begin{align*}
&\nabla_{\mathbf{x}} \mathbf{g}(t,\mathbf{x})|_{t, \mathbf{x}^{p}_{t}}(\mathbf{x}_{t}-\mathbf{x}^{p}_{t})
\\=&\nabla_{\mathbf{x}} \mathbf{f}(\mathbf{x},\mathbf{u}^{p}_{t}-\mathbf{L}_{t}(\mathbf{x}-\mathbf{x}^{p}_{t}))|_{ \mathbf{x}^{p}_{t} }(\mathbf{x}_{t}-\mathbf{x}^{p}_{t})
\\=&\nabla_{\mathbf{x}} \mathbf{f}(\mathbf{x},\mathbf{u})|_{ \mathbf{x}^{p}_{t},\mathbf{u}^{p}_{t}-\mathbf{L}_{t}(\mathbf{x}-\mathbf{x}^{p}_{t}) }(\mathbf{x}_{t}-\mathbf{x}^{p}_{t})
\\&+\nabla_{\mathbf{u}} \mathbf{f}(\mathbf{x},\mathbf{u})|_{ \mathbf{x}^{p}_{t},\mathbf{u}^{p}_{t}-\mathbf{L}_{t}(\mathbf{x}-\mathbf{x}^{p}_{t})}
\\&~~~~~~~~~~~~~~~\times\frac{\partial(\mathbf{u}^{p}_{t}-\mathbf{L}_{t}(\mathbf{x}-\mathbf{x}^{p}_{t}))}{\partial\mathbf{x}}|_{ \mathbf{x}^{p}_{t}}(\mathbf{x}_{t}-\mathbf{x}^{p}_{t})
\\=&\nabla_{\mathbf{x}} \mathbf{f}(\mathbf{x},\mathbf{u})|_{ \mathbf{x}^{p}_{t},\mathbf{u}^{p}_{t} }(\mathbf{x}_{t}-\mathbf{x}^{p}_{t})
\\&+\nabla_{\mathbf{u}} \mathbf{f}(\mathbf{x},\mathbf{u})|_{ \mathbf{x}^{p}_{t},\mathbf{u}^{p}_{t} }(-\mathbf{L}_{t})(\mathbf{x}_{t}-\mathbf{x}^{p}_{t})
\\=&\mathbf{A}_{t}(\mathbf{x}_{t}-\mathbf{x}^{p}_{t})
+\mathbf{B}_{t}(-\mathbf{L}_{t})(\mathbf{x}_{t}-\mathbf{x}^{p}_{t})
=\mathbf{D}_{t}(\mathbf{x}_{t}-\mathbf{x}^{p}_{t}).
\end{align*}
Therefore,
\begin{align}
\mathbf{g}(t,\mathbf{x}_{t})=&\mathbf{D}_{t}(\mathbf{x}_{t}-\mathbf{x}^{p}_{t})
+\boldsymbol{\omega}_{t}
+o(e^{\mathbf{x}}_{t}),\text{   as } e^{\mathbf{x},\boldsymbol{\omega}}_{t}\rightarrow 0.
\end{align}

\textit{Validity of the linearization:} Let us analyze the validity of \eqref{eq:linearized system fully observed} using the Wentzell-Freidlin theory discussed in Section \ref{sec:Small Random Perturbations of a Non-Linear System }. Let us assume that the noise process is $ \boldsymbol{\omega}_t=\epsilon\boldsymbol{w}_t $, where $ \boldsymbol{w}_t $ is a Wiener process as described in Section \ref{sec:Small Random Perturbations of a Non-Linear System }, and $ \epsilon>0 $. Now, for a time-varying system, the probability that the error $ |\!|\tilde{\mathbf{x}}_t|\!| $ is less than a given $ \delta>0 $ can be calculated using large deviations theory. In particular, the discussion in Section \ref{sec:Small Random Perturbations of a Non-Linear System } holds for process $ \mathbf{g} $. However, we require the function $ \mathbf{g} $ to satisfy a uniform Lipschitz continuity condition, for which uniform Lipschitz continuity of process model $ \mathbf{f} $ is sufficient. This is because, if
\begin{align*}
|\!|\mathbf{f}(\mathbf{x}_1, \mathbf{u}_1)&-\mathbf{f}( \mathbf{x}_2, \mathbf{u}_2)|\!|
\le K_f(|\!|\mathbf{x}_1-\mathbf{x}_2|\!|\!+\!|\!|\mathbf{u}_1-\mathbf{u}_2|\!|),
\end{align*}
where $ \mathbf{x}_1, \mathbf{x}_2\in \mathbb{R}^{n_x} $, and $ \mathbf{u}_1, \mathbf{u}_2\in \mathbb{R}^{n_u} $, in addition to smoothness of the nominal trajectory (which is calculated as in \eqref{eq:p-traj}) on the interval $ [0, K] $, and we have the Lipschitz continuity of $ \mathbf{g} $, as well.

\textit{Effect of feedback on the linearization error:} Note that before applying the feedback law, equation \eqref{eq:non-linear system equations fully observed} depends on both $ \mathbf{u} $ and $ \boldsymbol{\omega} $. The influence of $ \boldsymbol{\omega} $ can be analyzed using large deviations theory; however, it is the feedback law that limits the error of linearization caused by the control actions and converts the control action error into the state error. Moreover, the feedback effectively changes the drift term of the diffusion process and affects the validity region's probability through the action functional.

\subsection{Main Results}
In this section, we quantify the overall performance obtained from the separated design. The proofs are provided in the appendix.

\begin{lemma}\textup{\textbf{State Error Propagation:}}\label{lemma:State fully observed} Let $ \boldsymbol{\omega}_t=\epsilon\boldsymbol{w}_t $, where $ \boldsymbol{w}_t $ is a Gaussian process as described in section \ref{sec:Small Random Perturbations of a Non-Linear System }, and $ \epsilon>0 $. Let the state error be $ \tilde{\mathbf{x}}_{t}=\mathbf{x}_{t}-\mathbf{x}^{p}_{t} $ for $ t\ge 0 $. Then, for $ t\ge 0 $ the non-recursive state error propagation, $ \tilde{\mathbf{x}}_{t+1} $, in terms of the independent variables, including process noise at each time step can be written as follows:
\begin{align}\label{eq:State Error Propagation fully observed}
\tilde{\mathbf{x}}_{t+1}=\sum\limits_{s=0}^{t}\tilde{\mathbf{D}}^{\boldsymbol{\omega}}_{s,t}\boldsymbol{\omega}_s+o(\delta),
\text{   as } \epsilon\rightarrow 0,
\end{align}
where we have:
\begin{itemize}\compresslist
\item $ \mathbf{D}_{0}:=\mathbf{A}_0 $
\item $ \tilde{\mathbf{D}}_{t_1:t_2} = \Pi_{t=t_1}^{t_2}\mathbf{D}_{t}, t_2\ge t_1\ge 0 $, otherwise, it is the identity matrix;
\item $ \tilde{\mathbf{D}}^{\boldsymbol{\omega}}_{s,t}:=\tilde{\mathbf{D}}_{s+1:t}, 0\le s\le t-1, t\ge 1 $; and
\item $ \tilde{\mathbf{D}}^{\boldsymbol{\omega}}_{t,t}:=\tilde{\mathbf{D}}_{t+1:t}=\mathbf{I}, t\ge 0 $.
\end{itemize}
\end{lemma}

The following lemma follows directly by taking into account the feedback law in the result of Lemma \ref{lemma:State fully observed}.
\begin{lemma}\textup{\textbf{Control Error Propagation:}}\label{lemma:Control fully observed} Let $ \boldsymbol{\omega}_t=\epsilon\boldsymbol{w}_t $, where $ \boldsymbol{w}_t $ is a Gaussian process as described in section \ref{sec:Small Random Perturbations of a Non-Linear System }, and $ \epsilon>0 $. Let the control error be $ \tilde{\mathbf{u}}_{t}=\mathbf{u}_{t}-\mathbf{u}^{p}_{t} $ for $ t\ge 0 $. Then, for $ t\ge 0 $ the non-recursive control error propagation, $ \tilde{\mathbf{u}}_{t+1} $, in terms of the independent variables, including process noise at each time step can be written as follows:
\begin{align*}
\tilde{\mathbf{u}}_{t+1}=-\sum\limits_{s=0}^{t}\mathbf{L}^{\boldsymbol{\omega}}_{s,t+1}\boldsymbol{\omega}_s+o(\delta),
\text{   as } \epsilon\rightarrow 0,
\end{align*}
where $ \mathbf{L}^{\boldsymbol{\omega}}_{s,t+1}:=\mathbf{L}_{t+1}\tilde{\mathbf{D}}^{\boldsymbol{\omega}}_{s,t}, t\ge 0, t\ge s\ge 0 $. Moreover, the validity region of the above equation is the same as for \eqref{eq:State Error Propagation fully observed} in Lemma \ref{lemma:State fully observed}.
\end{lemma}

Next, we linearize of the cost function and provide the separation result for a fully observed system.

\textit{Linearization of the cost function:} Using the Taylor approximation around the nominal trajectories of state and control actions yields
\begin{align}
J=J^{p}+\sum_{t=0}^{K-1}(\mathbf{C}^{\mathbf{x}}_t\tilde{\mathbf{x}}_t+ \mathbf{C}^{\mathbf{u}}_t\tilde{\mathbf{u}}_t)+ \mathbf{C}^{\mathbf{x}}_K\tilde{\mathbf{x}}_K+o(e^{\mathbf{x},\mathbf{u}}_{\tilde{J}_{1}}),
\end{align}
where we assume that the cost function is continuously differentiable. Moreover:
\begin{itemize}\compresslist
\item $ J^{p}:=\sum_{t=0}^{K-1}c_t(\mathbf{x}^{p}_t,\mathbf{u}^{p}_t)+c_K(\mathbf{x}^{p}_K) $ denotes the nominal cost;
\item $ J_{1}:=J^{p} + \sum_{t=0}^{K-1}(\mathbf{C}^{\mathbf{x}}_t\tilde{\mathbf{x}}_t+ \mathbf{C}^{\mathbf{u}}_t\tilde{\mathbf{u}}_t)+ \mathbf{C}^{\mathbf{x}}_K\tilde{\mathbf{x}}_K $ is the first order approximation of the cost function;
\item $ \tilde{J}_{1}:= \sum_{t=0}^{K-1}(\mathbf{C}^{\mathbf{x}}_t\tilde{\mathbf{x}}_t+ \mathbf{C}^{\mathbf{u}}_t\tilde{\mathbf{u}}_t)+ \mathbf{C}^{\mathbf{x}}_K\tilde{\mathbf{x}}_K $ is the first order error in the cost by our approximation scheme. Therefore, $ \tilde{J}_{1}=J_{1}-J^{p} $;
\item $ \mathbf{C}^{\mathbf{x}}_t=\nabla_{\mathbf{x}} c_t(\mathbf{x},\mathbf{u})|_{\mathbf{x}^{p}_{t}, \mathbf{u}^{p}_{t}} $;
\item $ \mathbf{C}^{\mathbf{u}}_t=\nabla_{\mathbf{u}} c_t(\mathbf{x},\mathbf{u})|_{\mathbf{x}^{p}_{t}, \mathbf{u}^{p}_{t}} $;
\item $ \mathbf{C}^{\mathbf{x}}_K=\nabla_{\mathbf{x}} c_K(\mathbf{x})|_{\mathbf{x}^{p}_{K}} $; and
\item $ e^{\mathbf{x},\mathbf{u}}_{\tilde{J}_{1}}:=\sum_{t=1}^{K-1}(|\!|\tilde{\mathbf{x}}_t|\!|+|\!|\tilde{\mathbf{u}}_t|\!|)+|\!|\tilde{\mathbf{x}}_K|\!| $ is the linearization error.
\end{itemize}
Note that since the error term is in terms of state and control at all time steps, the probability of this equation holding true is equivalent to the probability of the latest time-step term still being in the vicinity of the nominal trajectory at that step. Therefore, the probability that this last equation is valid can be calculated as the probability that $ |\!|\tilde{\mathbf{x}}_K|\!|\ge\delta $ for $ \delta>0 $, which is given by equation \eqref{eq:asymptotics D_t 1 } for process $ \mathbf{g} $ defined in equation \eqref{eq:g function} and using $ \mathbb{D}_K=\mathrm{cl}(\mathbb{B}^{c}_{\delta}(\mathbf{x}^{p}_{K})) $ in Theorem \ref{theorem: Rate of Convergence }. As a result, all the previous steps will remain within the same tube around the nominal trajectory and the total error will still be of the order of $ \delta $. Therefore, given this probability, we have:
\begin{align}\label{eq:first order cost with delta}
J= J^{p}+\sum_{t=0}^{K-1}(\mathbf{C}^{\mathbf{x}}_t\tilde{\mathbf{x}}_t+ \mathbf{C}^{\mathbf{u}}_t\tilde{\mathbf{u}}_t)+ \mathbf{C}^{\mathbf{x}}_K\tilde{\mathbf{x}}_K+o(\delta),
\end{align}
$ \epsilon\rightarrow 0 $. Hence, $ J-J_{1}=o(\delta) $ as $ \epsilon\rightarrow0 $ with probability given in equation \eqref{eq:asymptotics D_t 1 } for $ t=K $.

Next, we provide the main result regarding the expected first order error of the cost function.
\begin{theorem}\textup{\textbf{First Order Cost Function Error:}}\label{theroem:First Order Cost Function Error (Fully Observed Case)} Let us denote the first order cost function error by $ \tilde{J}_{1} $. Given that process noises are zero mean i.i.d., under a first-order approximation for the small noise paradigm, the stochastic cost function is dominated by the nominal part of the cost function. Moreover the expected first-order error is zero. That is,
\begin{align*}
\mathbb{E}[\tilde{J}_{1}]=0.
\end{align*}
Moreover, if the process noise at each time step is distributed according to a zero mean Gaussian distribution, then $ \tilde{J}_{1} $ also has a zero mean Gaussian distribution.
\end{theorem}
The above result says that the random perturbation in the stochastic running cost form the nominal is zero mean if the linearization holds. From Wentzell-Freidlin theory, we have already established that the linearization holds with a probability exponentially close to 1 as $\epsilon \rightarrow 0$. Hence, this implies that the expected stochastic cost is equal to the nominal cost with a very high probability as $\epsilon \rightarrow 0$. Therefore, it follows that the open loop nominal design can be done separately from the closed loop design, summarized bellow:
\begin{coroll}\textup{\textbf{Separation of the Closed Loop and Open Design Under Small Noise}}\label{coroll 1:Cost first order fully observed}
Based on Theorem \ref{theroem:First Order Cost Function Error (Fully Observed Case)}, under the small noise paradigm, as $ \epsilon\rightarrow0 $, the design of the feedback law can be done separately from the design of the open loop optimized trajectory. Furthermore, this result holds with a probability that exponentially tends to one as $ \epsilon\rightarrow0 $.
\end{coroll}
\textit{Remark:} This result means that under a small noise assumption and assuming the existence of a feedback law (with LTV gain, which is designed separately), the open loop nominal trajectory of the system can be designed by replacing the stochastic equations with their nominal counterparts. This design tends to the optimal design with probability one (for the general class of Gaussian processes that are considered) as the intensity of noise tends to zero.

\textit{Remark:} It should be mentioned that while our general problem definition has only the process model as dynamics, other constraints on state or control can be considered as long as they share the same smoothness properties as the cost function.

\textit{Remark:} It is worth mentioning that although we have considered diffusion processes with additive white Gaussian noise, the theory in fact holds for a larger class of problems. On can appeal to more general results in \cite{wentzell2012limit} for time-inhomogeneous diffusion processes with non-additive white noise. In such cases, the action functional is usually calculated through the Legendre transform.

\textit{Remark:} As mentioned before, although we proved the results of this section for discrete time systems, one can prove the continuous-time versions of our results. This can be done, for instance, by reducing the sampling time and limiting it to zero, while utilizing results such as Fubini's theorem along with the similar conditional expectation theorem on It\^{o}'s stochastic integrals to exchange the integrations with the expectation. It should be mentioned that there also exists a discrete-time counterpart of the Wentzell-Freidlin theory as provided in \cite{wentzell2012limit}.

\textit{Remark:} Higher order designs and analysis of the cost function (or even the dynamics) are possible using a similar approach provided in this paper.

\textit{Remark:} In Ref. \cite{fleming1971stochastic}, for a special case of nonlinear systems where the process model is linear in the control variable, i.e., $ \mathbf{f}(\mathbf{x}_{t},\mathbf{u}_{t})= f_1(\mathbf{x}_{t}) + f_2(\mathbf{x}_{t})\mathbf{u}_{t} $, three results are proven. The first result, concerns the $ \epsilon $-optimality of the optimal deterministic law under convexity of $ J $ in the control (i.e., $ \mathbf{v}^{T}(\nabla_{\mathbf{u},\mathbf{u}} J)\mathbf{v}\succeq 0~, \forall \mathbf{v}$), and additional smoothness and regularity conditions. The second result concerns the $ \epsilon^{2} $-optimality of the optimal deterministic law under a stronger convexity condition of $ J $ in the control (i.e., $ \mathbf{v}^{T}(\nabla_{\mathbf{u},\mathbf{u}} J)\mathbf{v}\succeq c(|\!|\mathbf{u}|\!|)|\!|\mathbf{v}|\!|^{2}~, \forall \mathbf{v}$, $ c(\cdot):\mathbb{R}\rightarrow\mathbb{R} $ is a monotonically non-increasing positive function), and some smoothness and regularity conditions. The third result concerns the $ \epsilon $-optimality of the optimal deterministic sequence under the latter condition.
Our result, on the other hand, provides the $ \epsilon $-optimality of the proposed design approach for a broader class of processes $ \mathbf{f}(\mathbf{x}_{t},\mathbf{u}_{t}) $ with nonlinear dependence in the control variable and more general cost functions (most importantly, does not assume the linear dependence on the control sequence). In fact, our simulations are performed for a car-like robot with nonlinear dependence on the control variables.
\section{T-LQR: Trajectory-optimized LQR}\label{sec:T-LQR: Trajectory-optimized LQR}
In this section, we provide a design scheme based on the theory provided in the previous sections. This approach aims at designing an LQR controller with an optimal nominal underlying trajectory based on the separation result of Corollary \ref{coroll 1:Cost first order fully observed} and Theorem \ref{theroem:First Order Cost Function Error (Fully Observed Case)}. As a result, we term this method as the Trajectory-optimized LQR (T-LQR).

\begin{problem}\label{problem:Planning Problem 4}\textup{\textbf{Trajectory Planning Problem:}} Solve for the optimal trajectory:
\begin{subequations}
\begin{align}
\nonumber \min_{\mathbf{u}^{p}_{0:K-1}}~\sum\limits_{t=0}^{K-1}&c(\mathbf{x}^{p}_{t},\mathbf{u}^{p}_{t})+c_K(\mathbf{x}^{p}_{K})
\\s.t.~~\mathbf{x}^{p}_{t+1}&=\mathbf{f}(\mathbf{x}^{p}_t, \mathbf{u}^{p}_t), ~0\!\le\! t\! \le\! K\!-\!1,
\\\mathbf{x}^{p}_{0} &= \mathbf{x}_{0}.
\end{align}
\end{subequations}
\end{problem}

\textit{Optimized nominal trajectory:} Problem \ref{problem:Planning Problem 4} is a deterministic problem aiming for the best nominal performance. This problem utilizes the first order approximation of the cost function and optimizes the underlying nominal trajectory used in the design of the feedback law. We will denote the resulting optimized nominal trajectory of problem \ref{problem:Planning Problem 4} by $ \{\mathbf{x}^{o}_{t}\}_{t=0}^{K} $, $ \{ \mathbf{u}^{o}_{t}\}_{t=0}^{K-1}  $.

\textit{Feedback control:} The resulting trajectory from the optimization problem is optimized in terms of control effort and other constraints, such as a terminal constraint. Now, using the separation result, an LQR controller is designed to track the optimized nominal trajectory. Therefore, the LQR cost is designed for the tracking error $ \mathbf{x}_t-\mathbf{x}^{o}_t $. The resulting control policy is a feedback policy with LTV gain, and the evolution of $ \mathbf{x}_t $ is obtained from the original equation of the process model during the execution. Although we utilize an LQR controller, it is important to note that the separation result only assumes a linear form of feedback and other types of designs \cite{kumar2014control} can be used as well.

\textit{Linearization of system equations:} For simplicity, we denote the Jacobian matrices and every other variable associated with the optimized nominal trajectory with a superscript $ o $. The Jacobians are $ \mathbf{A}^{o}_t=\nabla_{\mathbf{x}} \mathbf{f}(\mathbf{x},\mathbf{u})|_{ \mathbf{x}^{o}_{t}, \mathbf{u}^{o}_{t} } $, and $ \mathbf{B}^{o}_t=\nabla_{\mathbf{u}} \mathbf{f}(\mathbf{x},\mathbf{u})|_{ \mathbf{x}^{o}_{t}, \mathbf{u}^{o}_{t}} $.

\begin{problem}\label{problem 5}\textup{\textbf{LQR Problem:}} Given the optimized nominal trajectory as
$ \{\mathbf{x}^{o}_{t}\}_{t=0}^{K} $ and $ \{ \mathbf{u}^{o}_{t}\}_{t=0}^{K-1}  $, and a planning horizon of $ K>0 $, solve the following LQR problem to track the nominal trajectory:
\begin{align}
\nonumber \min_{\mathbf{u}^{p}_{0:K-1}}
\sum_{t=1}^{K}[&(\mathbf{x}_t-\mathbf{x}^{o}_t)^T\mathbf{W}^{x}_{t}(\mathbf{x}_t-\mathbf{x}^{o}_t)+(\tilde{\mathbf{u}}^{o}_{t-1})^{T}\mathbf{W}^{u}_{t}\tilde{\mathbf{u}}^{o}_{t-1}]
\\s.t.~~\mathbf{x}^{o}_{t+1}&=\mathbf{A}^{o}_t\mathbf{x}^{o}_{t}+\mathbf{B}^{o}_t\mathbf{u}^{o}_{t}, ~0\!\le\! t\! \le\! K\!-\!1
\end{align}
where $ \tilde{\mathbf{u}}^{o}_t = \mathbf{u}_t-\mathbf{u}^{o}_t $ and $ \mathbf{W}^{u}_{t}, \mathbf{W}^{x}_{t}\succeq 0 $ are positive-definite matrices.
\end{problem}

\textit{Control policy:} The resulting control policy of problem \ref{problem 5} is a feedback policy as follows \cite{Kumar-book-86}:
\begin{align*}
\tilde{\mathbf{u}}^{o}_{t}=-\mathbf{L}^{o}_{t}(\mathbf{x}_t-\mathbf{x}^{o}_t),
\end{align*}
where the linear feedback gain $ \mathbf{L}^{o}_{t} $ is:
\begin{align*}
\mathbf{L}^{o}_{t} = (\mathbf{W}^{u}_{t}+(\mathbf{B}^{o}_{t})^T\mathbf{P}^{f}_{t+1}\mathbf{B}^{o}_{t})^{-1}(\mathbf{B}^{o}_{t})^T\mathbf{P}^{f}_{t+1}\mathbf{A}^{o}_{t},
\end{align*}
and the matrix $ \mathbf{P}^{f}_{t} $ is the result of backward iteration of the dynamic Riccati equation
\begin{align*}
&\mathbf{P}^{f}_{t-1} = (\mathbf{A}^{o}_{t})^T\mathbf{P}^{f}_{t}\mathbf{A}^{o}_{t}
\\&-\!(\mathbf{A}^{o}_{t})^T\mathbf{P}^{f}_{t}\mathbf{B}^{o}_{t}(\mathbf{W}^{u}_{t}+(\mathbf{B}^{o}_{t})^T\mathbf{P}^{f}_{t}\mathbf{B}^{o}_{t})^{-1}(\mathbf{B}^{o}_{t})^T\mathbf{P}^{f}_{t}\mathbf{A}^{o}_{t}\!+\!\mathbf{W}^{x}_{t},
\end{align*}
which is solvable with a terminal condition $ \mathbf{P}^{f}_{K}=\mathbf{W}^{x}_{t} $.
\begin{figure}[!]
\centering
   \subfloat[Optimized trajectory of problem \ref{problem:Planning Problem 4}.\label{fig:Optimized trajectory}]{\includegraphics[width=0.473\linewidth]{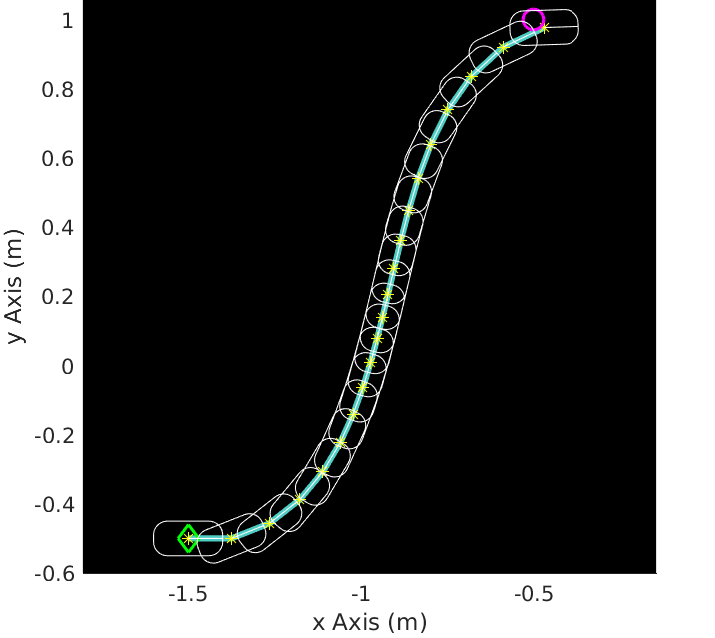}}
   \subfloat[A typical ground truth trajectory with noise standard deviation equal to 10\% of the maximum control signal.\label{fig:10 percent noise}]{\includegraphics[width=0.5\linewidth]{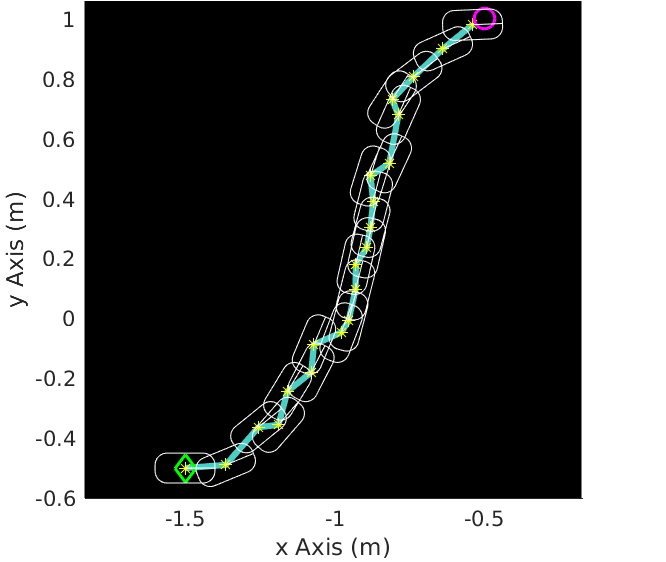}}
   \caption{Optimized vs. a typical execution trajectory for a car-like robot.\label{fig:trajectories}}\vspace{-4pt}
\end{figure}
\textit{Remark:} The computations involved in problem \ref{problem:Planning Problem 4} is of the order of $ O(Kn_{x}^{2}) $ for typically smooth dynamics for one iteration. Let us assume $ O(\ell) $ is the order of the number of iterations in the optimizer until convergence. The LQR policy calculation is of order of $ O(Kn_{x}^{3}) $. Therefore, overall, the design approach based on the separation principle of Corollary \ref{coroll 1:Cost first order fully observed} is $ O(\ell Kn_{x}^{2}+ Kn_{x}^{3}) $ for a typical process model (such as our example in the next section). The low computational complexity of this approach results in fast replanning in case of deviations during execution. This renders the first scheme to be eminently implementable for implementation in on-line applications.

\textit{Remark:} For the specific class of problems considered in \cite{fleming1971stochastic} (see the last remark in Section \ref{sec:Separation of Open Loop and Closed Loop Designs: Fully Observed Systems}) the design approach of \cite{fleming1971stochastic} requires calculation of the optimal control law through intractable dynamic programming. In contrast, the proposed design approach in this paper utilizes the tractable solution of Maximum Principle problem followed by an LQR design. Even implementing the result of \cite{fleming1971stochastic} through a model predictive approach would require more computations of at least an order of the planning horizon (from $ O(K) $ to $ O(K^{2}) $). In such an implementation, the online computations of the approach of \cite{fleming1971stochastic} require $ O(\ell Kn_{x}^{2}) $ calculations compared to only $ O(n_x^{2}) $ calculations in our algorithm.

\section{Example}\label{sec:Example}
Let us consider a car-like four-wheel robot with process model \cite{LaValle06}:
\begin{align}
\dot{x}=v\cos(\theta),~ \dot{y}=v\sin(\theta),~ \dot{\theta}=\frac{v}{L}\tan(\phi),
\end{align}
where $ (x,y,\theta) $ is the state, and $ (v, \phi) $ is the control input. We suppose that, $ |\phi|< \phi_{\max}=\pi/2 $, $ |v|\le v_{\max}=0.6 $, $ \mathbf{x}_0 = (-1.5, 0.5, 0) $, $ K=20 $, and the time discretization period is $ 0.7 $. We incorporate the control constraints and the terminal goal, $ \mathbf{x}_g = (-0.5, 1, 0) $, in the cost function. Last, the initial control sequence used for the optimization is just a sequence of zero inputs. The process noise is additive mean zero Gaussian noise with a standard deviation equal to $ \epsilon\max_t\{|\!|\mathbf{u}_t|\!|_{2}\} $. Figure \ref{fig:Optimized trajectory} shows the result of the optimization problem \ref{problem:Planning Problem 4} whereas Fig. \ref{fig:10 percent noise} shows a typical ground truth trajectory with $ \epsilon=0.1 $. We have used MATLAB 2016b and its $ \mathtt{fmincon} $ solver for simulations.
\begin{figure}[!]
\centering
   \subfloat[Feedback-compensated system.\label{fig:NMSE evolution}]{\includegraphics[width=0.5\linewidth]{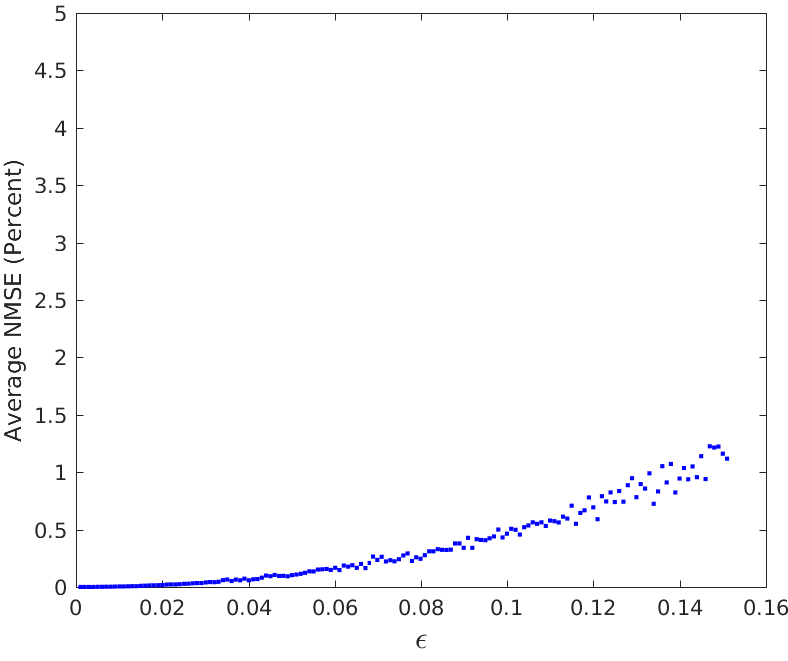}}
   \subfloat[Open-loop system.\label{fig:NMSE evolution open loop}]{\includegraphics[width=0.5\linewidth]{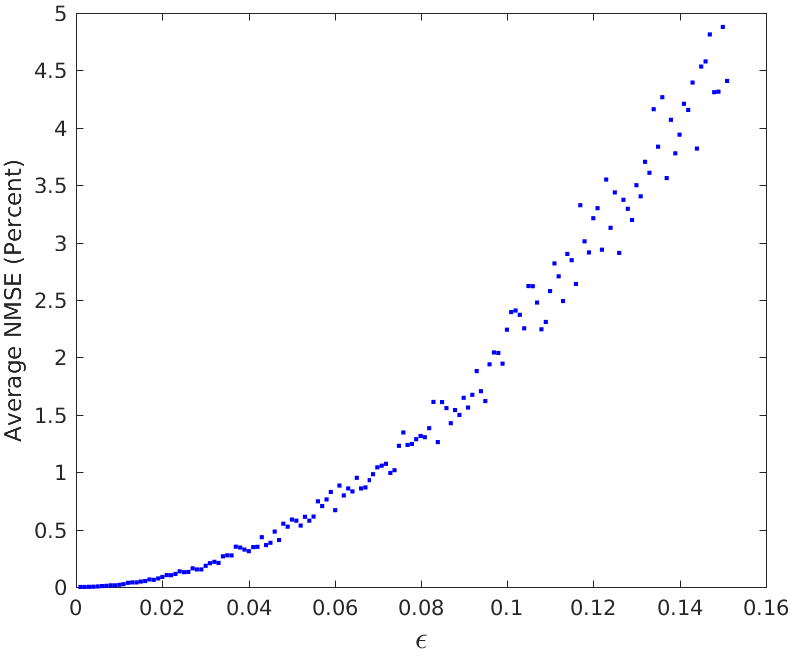}}
   \caption{Evolution of average NMSE as $ \epsilon\downarrow0 $ for a feedback compensated and open loop system with the same nominal trajectories.\label{fig:NMSE evolution both}}\vspace{-4pt}
\end{figure}

In the next experiment, we increase $ \epsilon $ from 0.001 to 0.1501, in step sizes of 0.001. For each value of $ \epsilon $, we execute the resulting policy 100 times and compute the average Normalized Mean Squared Error (NMSE) as:
\begin{align}
\mathrm{Average\text{ }NMSE}\text{ }(\%) = \frac{1}{100}\sum_{j=1}^{100}\frac{|\!|\mathbf{x}^{p}-\mathbf{x}^{j}|\!|_{2}^{2}}{|\!|\mathbf{x}^{p}|\!|_{2}^{2}}\times 100,
\end{align}
where $ \mathbf{x}^{p} $ indicates the planned trajectory and $ \mathbf{x}^{j} $ indicates the ground truth trajectory at $ j $th experiment. The results of this experiment are shown in Fig. \ref{fig:NMSE evolution}, where the evolution of the average NMSE is depicted for various values of noise level $ \epsilon $. As indicated in this figure, as $ \epsilon\downarrow0 $, the average NMSE tends to zero at an exponential rate, which is consistent with the theory developed in Section \ref{sec:Small Random Perturbations of a Non-Linear System }. Moreover, this figure indicates that through the feedback compensation, moderate noise levels can be tolerated, rather than just small levels.

Last, Fig. \ref{fig:NMSE evolution open loop} depicts the evolution of the average NMSE for an experiment with the same setting as in Fig. \ref{fig:NMSE evolution}, except that only the open-loop planned control sequence is applied during execution. As predicted by the theory, the error still decreases exponentially as the noise level decreases. However, the rate of convergence is about one-fifth of the previous rate. The results of Fig. \ref{fig:NMSE evolution both} show that our design can be used for relatively moderate levels of noise, using the power of feedback.

\textit{Remark:} In practice, if at any point in the execution the calculated error exceeds a threshold, very rapid replanning can be triggered very fast due to the low computational burden of the optimization problem.

\section{Conclusion}\label{sec:Conclusion} 
We have presented a design approach that separates the design of the open-loop nominal trajectory and the closed-loop feedback policy for fully-observed nonlinear stochastic systems with Gaussian distributions. We have shown that under a small-noise assumption, the stochastic cost function is dominated by the nominal part of the cost function and the expected first order linearization error is of mean zero. This results in a reliable rapid planning method that is provably near-optimal. It can be used in robotic path planning and control, and potentially in other applications.

\bibliographystyle{IEEEtran}
\bibliography{AliAgha}
\appendix\vspace{-3pt}
\section{Proof of Results}\label{appdx:Proof of the Results For Fully Observed Problem}
\begin{proof}\textup{\textbf{Lemma \ref{lemma:State fully observed}: State Error Propagation}}

Ignoring the validity region,
\begin{align*}
\tilde{\mathbf{x}}_{t+1}\!\!&=\!\!\mathbf{A}_t\tilde{\mathbf{x}}_t + \mathbf{B}_t\tilde{\mathbf{u}}_t +\boldsymbol{\omega}_t= (\mathbf{A}_t - \mathbf{B}_t\mathbf{L}_t)\tilde{\mathbf{x}}_t +\boldsymbol{\omega}_t
\\&=:\! \mathbf{D}_t\tilde{\mathbf{x}}_t \!+\!\boldsymbol{\omega}_t
=:\!\!\tilde{\mathbf{D}}_{0:t}\tilde{\mathbf{x}}_0\!+\!\sum\limits_{r=0}^{t}\tilde{\mathbf{D}}_{r+1:t}\boldsymbol{\omega}_r
=:\!\!\sum\limits_{s=0}^{t}\tilde{\mathbf{D}}^{\boldsymbol{\omega}}_{s,t}\boldsymbol{\omega}_s.
\end{align*}
Note that using the definition of $ \tilde{\mathbf{x}}_{t} $, the initial state error is $ \tilde{\mathbf{x}}_{0} = \mathbf{x}_0-\mathbf{x}^{p}_0=\mathbf{x}_0-\mathbf{x}_0=\mathbf{0} $. Likewise, the state error at time-step $ 1 $ is $ \tilde{\mathbf{x}}_{1}=\mathbf{A}_0\tilde{\mathbf{x}}_0 +\boldsymbol{\omega}_0=\boldsymbol{\omega}_0 $. Moreover, these errors are consistent with the lemma using the definitions provided and the indicator function notation.

Now, since this equation utilizes the linearizations at all steps, its error is within $ o(\delta) $, if $ |\!|\tilde{\mathbf{x}}_{s}|\!|\le \delta $ for all $ s\le t $. Moreover, the probability that equation \eqref{eq:State Error Propagation fully observed} is valid (i.e., the linearizations are valid with $ o(\delta) $ error for the entire trajectory up to time $ t $) is the same as the probability that the linearization is valid on the last step (i.e., step $ t $). This is due to Wentzell-Freidlin theory. Now, the probability that $ |\!|\tilde{\mathbf{x}}_{t}|\!|\ge \delta $ is given by \eqref{eq:asymptotics D_t 1 } for process $ \mathbf{g} $ defined in \eqref{eq:g function}, and $ \mathbb{D}_t=\mathrm{cl}(\mathbb{B}_{\delta}(\mathbf{x}^{p}_{t})) $ for Theorem \ref{theorem: Rate of Convergence }. Therefore, as $ \epsilon\rightarrow0 $, the probability of $ |\!|\mathbf{x}_{t}-\mathbf{x}^{p}_{t}|\!|\ge\delta $ is calculated as in equation \eqref{eq:asymptotics D_t 1 }, which tends exponentially to zero. Last, note that through Wentzell-Freidlin theory, the validity of linearization only depends on the aggregated effect of the random perturbations at steps prior to $ t $, and there is no need to individually bound the noise at each step.
\end{proof}

\begin{proof}\textup{\textbf{Lemma \ref{lemma:Control fully observed}, Control Error Propagation}}

\textit{Replacing state error in the control law:} Using the result of Lemma \ref{lemma:State fully observed}, we can rewrite $ \tilde{\mathbf{u}}_{t+1} $ for $ t\ge -1 $ as follows:
\begin{align*}
\nonumber\tilde{\mathbf{u}}_{t+1}\!=\!-\mathbf{L}_{t+1}\tilde{\mathbf{x}}_{t+1}
\!=\!-\mathbf{L}_{t+1}\sum\limits_{s=0}^{t}\tilde{\mathbf{D}}^{\boldsymbol{\omega}}_{s,t}\boldsymbol{\omega}_s
\!=:\!-\sum\limits_{s=0}^{t}\mathbf{L}^{\boldsymbol{\omega}}_{s,t+1}\boldsymbol{\omega}_s.
\end{align*}

Note that $ \tilde{\mathbf{u}}_{0}=\mathbf{0} $, and the last formula is consistent with this error using the definitions provided in the lemma.
\end{proof}

\begin{proof}\textup{\textbf{Theorem \ref{theroem:First Order Cost Function Error (Fully Observed Case)}, Cost Function Error}}

Using the linearization process described previously, we can write the cost function error as $ \mathbb{E}[\tilde{J_1}]= \mathbb{E}[\sum_{t=0}^{K-1} (\mathbf{C}^{\mathbf{x}}_t\tilde{\mathbf{x}}_t+ \mathbf{C}^{\mathbf{u}}_t\tilde{\mathbf{u}}_t)+ \mathbf{C}^{\mathbf{x}}_K\tilde{\mathbf{x}}_K ] $. Utilizing the assumption that the process noise is zero mean i.i.d., $ \mathbb{E}[\boldsymbol{\omega}_t]={0} $ for all $ t $. Moreover, $ \tilde{\mathbf{x}}_0=\mathbf{0} $ which follows from the fact that $ \mathbf{x}_0=\mathbf{x}^{p}_0 $. Therefore, using the linearity of the expectation operator and Lemmas \ref{lemma:State fully observed} and \ref{lemma:Control fully observed}, we can rewrite $ \mathbb{E}[\tilde{J_1}] $ as follows:
\begin{align*}
\mathbb{E}[\tilde{J_1}]\!\!=\!\!&\sum_{t=0}^{K-1} (\mathbf{C}^{\mathbf{x}}_t\mathbb{E}[\tilde{\mathbf{x}}_t]+ \mathbf{C}^{\mathbf{u}}_t\mathbb{E}[\tilde{\mathbf{u}}_t])+ \mathbf{C}^{\mathbf{x}}_K\mathbb{E}[\tilde{\mathbf{x}}_K]
\\=&\sum_{t=0}^{K-1} \mathbf{C}^{\mathbf{x}}_t\mathbb{E}[\sum\limits_{s=0}^{t-1}\tilde{\mathbf{D}}^{\boldsymbol{\omega}}_{s,t-1}\boldsymbol{\omega}_s]
\!\!+\!\!\sum_{t=0}^{K-1}\mathbf{C}^{\mathbf{u}}_t\mathbb{E}[
-\sum\limits_{s=0}^{t-1}\mathbf{L}^{\boldsymbol{\omega}}_{s,t}\boldsymbol{\omega}_s]
\\&+\mathbf{C}^{\mathbf{x}}_K\mathbb{E}[\sum\limits_{s=0}^{K-1}\tilde{\mathbf{D}}^{\boldsymbol{\omega}}_{s,K-1}\boldsymbol{\omega}_s]
\\=&\!\!\sum_{t=0}^{K\!-\!1}\!\sum\limits_{s=0}^{t\!-\!1} \!\!\mathbb{E}[\!(\mathbf{C}^{\mathbf{x}}_t\tilde{\mathbf{D}}^{\boldsymbol{\omega}}_{s,t\!-\!1}
\!\!-\!\!\mathbf{C}^{\mathbf{u}}_t\mathbf{L}^{\boldsymbol{\omega}}_{s,t})\boldsymbol{\omega}_s\!]
\!\!+\!\!\!\sum\limits_{s=0}^{K\!-\!1}\!\!\mathbb{E}[\!\mathbf{C}^{\mathbf{x}}_K\tilde{\mathbf{D}}^{\boldsymbol{\omega}}_{s,K\!-\!1}\boldsymbol{\omega}_s\!]
\\=:&\sum_{t=0}^{K}\sum\limits_{s=0}^{t-1} \mathbb{E}[(\mathbf{w}_{s,t})^{T}\boldsymbol{\omega}_s]
=\sum_{t=0}^{K}\sum\limits_{s=0}^{t-1}\sum_{j=1}^{n_u}w_{s,t}^{j}\mathbb{E}[\omega_s^{j}]
=0.
\end{align*}
where $ \mathbf{w}^{s,t}:=(\mathbf{C}^{\mathbf{x}}_t\tilde{\mathbf{D}}^{\boldsymbol{\omega}}_{s,t-1}
-\mathbf{C}^{\mathbf{u}}_t\mathbf{L}^{\boldsymbol{\omega}}_{s,t})^{T}, t-1\ge s\ge 0, K-1\ge t\ge 0 $, $ \mathbf{w}^{s,K}:=(\mathbf{C}^{\mathbf{x}}_K\tilde{\mathbf{D}}^{\boldsymbol{\omega}}_{s,K-1})^{T}, K-1\ge s\ge 0 $. Moreover, $ \mathbf{w}_{s,t}:=(w_{s,t}^{1}, \cdots, w_{s,t}^{n_u})^{T} $ is a vector of the same size of $ \boldsymbol{\omega}_s=(\omega_{s}^{1}, \cdots, \omega_{s}^{n_u})^{T} $.
\end{proof}
\end{document}